\documentclass[journal]{IEEEtran}
\IEEEoverridecommandlockouts

\usepackage[numbers,sort,compress,square]{natbib}
\usepackage{amsmath,amssymb,amsfonts}
\usepackage{algorithmic}
\usepackage{graphicx}
\usepackage{textcomp}
\usepackage{xcolor}
\usepackage{lipsum}
\usepackage{hyperref} %
\usepackage{multirow}
\usepackage{caption}

\newif\ifanonymous
\anonymousfalse %

\newcommand{\anon}[1]{\ifanonymous\else#1\fi}
\newcommand{\emoji}[1]{\includegraphics[height=1em]{#1}}

\newif\ifreviewchanges
\reviewchangesfalse    %

\newcommand{\review}[1]{\ifreviewchanges\textcolor{blue}{#1}\else#1\fi}

\begin{document}
\bstctlcite{IEEEexample:BSTcontrol}
\title{  \anon{\emoji{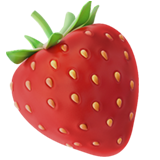}} \textbf{DexFruit}: Dexterous Manipulation and Gaussian Splatting Inspection of Fruit

\anon{
\thanks{$[\cdot]^{1}$ are with the Department of Mechanical Engineering,
$[\cdot]^{2}$ Department of Computer Science,
        Stanford University, Stanford CA, USA.
        Emails: \{swann, aqiu34, mastro1, azhang11, morstein, kairayle, monroek\}@stanford.edu}
\thanks{A. Swann is supported by NSF GRFP Fellowship No.
DGE-2146755. This work was also supported in part by NSF Grant No. 2142773 and 2220867}
\thanks{$^\ast$Both authors contributed equally to this work.}
}
}

\anon{
\author{Aiden Swann$^{1\ast}$, Alex Qiu$^{1\ast}$, Matthew Strong$^2$, Angelina Zhang$^1$, Samuel Morstein$^1$, \\ Kai Rayle$^1$, Monroe Kennedy III$^{1,2}$}
}

\maketitle

\begin{abstract}
DexFruit is a robotic manipulation framework that enables gentle, autonomous handling of fragile fruit and precise evaluation of damage. \review{Soft fruits have long faced an issue of produce loss in both the harvesting and post-harvesting processes due to their extreme fragility and susceptibility to bruising, making them one of the hardest produce type to manipulate with automation.} In this work, we demonstrate by using optical tactile sensing, autonomous manipulation of fruit with minimal damage can be achieved. We show that our tactile informed diffusion policies outperform baselines in both reduced bruising and pick-and-place success rate across three fruits: strawberries, tomatoes, and blackberries. In addition, we introduce \textbf{FruitSplat}, a novel technique to represent and quantify visual damage in a high-resolution 3D representation via 3D Gaussian Splatting (3DGS). Existing metrics for measuring damage lack quantitative rigor or require expensive equipment. With FruitSplat, we distill a 2D fruit mask as well as a 2D bruise segmentation mask into the 3DGS representation \review{from just a web-cam video}. Furthermore, this representation is modular and general, compatible with any relevant 2D model. Overall, we demonstrate a 92\% grasping policy success rate, up to a 15\% reduction in visual bruising, and up to a 31\% improvement in grasp success rate on challenging fruit compared to our baselines across our three tested fruits. We rigorously evaluate this result with over \textbf{630 trials}. \anon{Please checkout our website, which contains our code and datasets at \url{https://dex-fruit.github.io/.}}

\end{abstract}

\section{Introduction}
\label{sec:intro}

Demographic-driven labor shortages and a growing population present significant challenges for the global food supply \cite{calvin2022adjusting, luo2017us, godfray2010food}. \review{To address these impending issues, the agricultural industry has taken many strides into increased applications of machinery and automation \cite{edan2009automation, mahmud2020robotics}.} While some agricultural production is highly mechanized, such as grain and cereal, automation for fruits and vegetables is comparatively limited, often requiring workers to manually harvest produce \cite{calvin2022adjusting, umn_harvesting_strawberries}. Fruit production, in particular, requires a high level of labor-intensive operations to be performed, primarily due to the delicate handling required to preserve quality \cite{umn_harvesting_strawberries}. Fruits, such as strawberries, demand significant human labor because of their fragile structure and short post-harvest shelf life; they are typically hand-picked to minimize bruising and decay \cite{umn_harvesting_strawberries}. Although there are some mechanized solutions for fruit harvesting, they often do not meet the quality standards of the fresh market, damage fruit more than their human counterparts \cite{zhou2022intelligent}, and still involve significant human labor in sorting. Soft fruits in particular are especially challenging for robotic systems, as they require extra care and precision in harvesting and manipulation not required for other produce \cite{lenz2024hortibotadaptivemultiarmrobotic, gundertendondriven, myers_hand_harvest}\review{, presenting a clear need for dexterous robotic manipulation.} 

\begin{figure}
    \centering
    \includegraphics[width=0.95\linewidth]{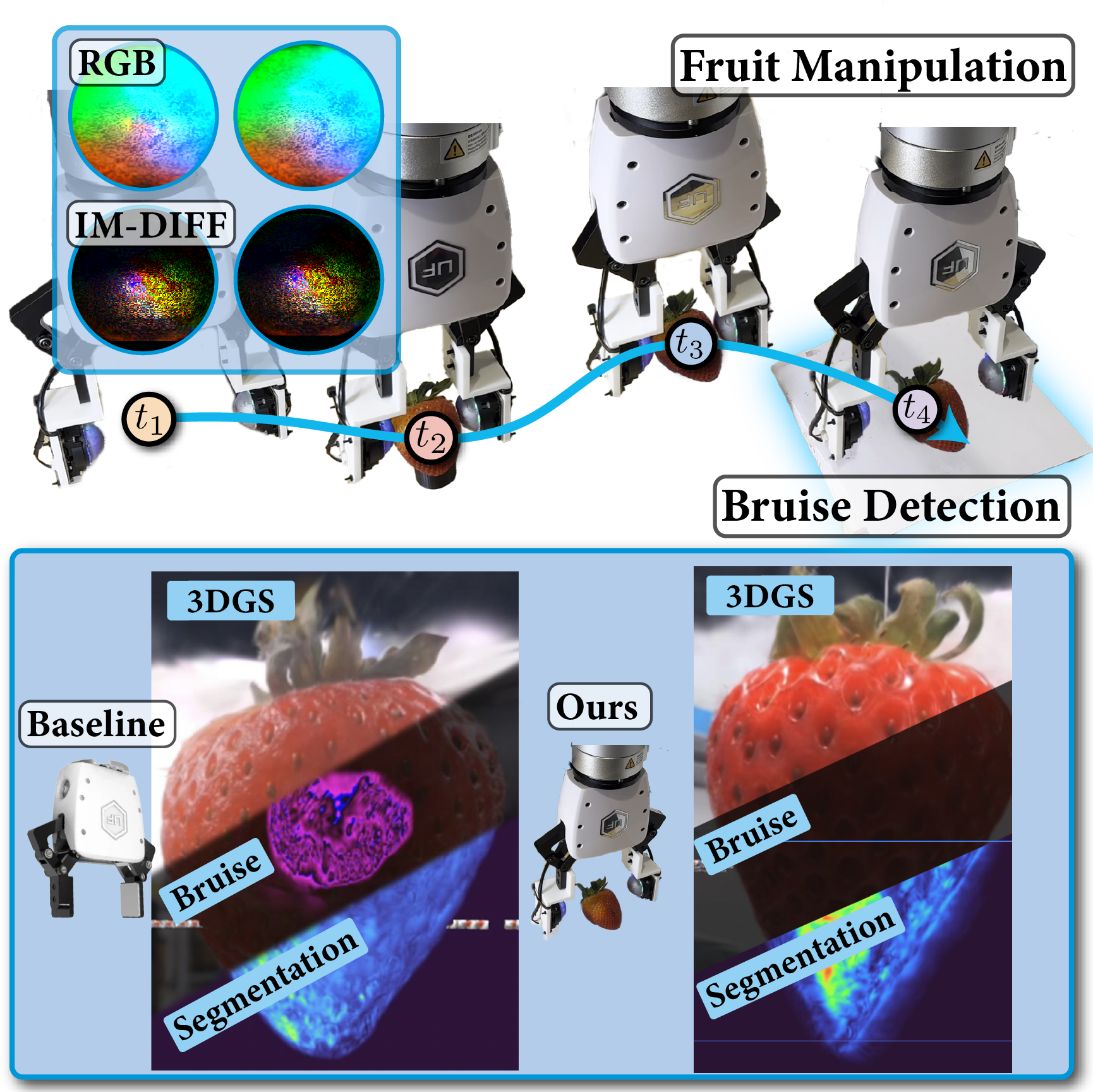}
    \caption{Our framework, DexFruit, enables safe manipulation of fragile fruit using optical tactile feedback. We further present FruitSplat, a method for accurate 3D reconstruction, automated segmentation, and bruise localization in strawberries. }
    \label{fig:splash}
\end{figure}

\review{Beyond harvesting, the industry has also turned to robotics and automation for the post-harvest process, which includes sorting, quality control, packing, and shipping. These processes incur substantial losses to farmers due to post-harvest damage, accounting for almost a third of all of the world's produce \cite{das2025post}. 
This significant loss of food compounded with high process inefficiencies and labor costs have specifically driven the industry to integrate robots into this post-harvesting downstream pipeline, applying soft-grippers and various computer vision techniques for manipulation and damage assessment \cite{singh2022recentpostharvestadvancement}. With automation introduced into the post-harvest process, pick-and-place operations have become the most repetitive robotic manipulation task \cite{quality-driven-post-harvest-2022}. In order for these pick-and-place tasks to be successful, the robot must maintain a stable but soft grasp on the target produce to ensure minimal damage. Recent advances in imitation learning have allowed for tasks like pick-and-place to be executed with ease, yet most of these methods employ rigid grippers and target rigid objects, neglecting damage mitigation. Moreover, these methods do not address the question of post-manipulation damage, a crucial step that must be tightly coupled with automated manipulation. As of now, there is a lack of low-cost, automated methods for assessing damage quantitatively for quality control. Practical damage metrics are a pre-requisite to real-world robot policy benchmarking and will facilitate continued development of soft fruit manipulation.}

\review{To address these gaps, we introduce \textbf{DexFruit}, a novel framework that integrates tactile-sensing into imitation learning for careful manipulation of soft fruit and present a stand-alone post-manipulation damage analysis method \textbf{FruitSplat}, which offers a quantitative, 3D solution for bruise detection. Our method employs DenseTact sensors \cite{doDT2.0} to manipulate strawberries, cherry tomatoes, and blackberries with minimal bruising or damage. We implement Diffusion Policy with tactile feedback from DenseTact sensors, which are mounted on simple general-purpose parallel jaw-grippers. FruitSplat is a 3D Gaussian Splatting method that integrates 2D bruising information from a fine-tuned YOLO \cite{yolo} segmentation model into a 3D representation, providing both a qualitative and quantitative measurement of post-manipulation damage. Using only a simple webcam, FruitSplat bridges the gap between high-fidelity scanning techniques and traditional deep learning methods. Our framework aims to address the crucial need for dexterous manipulation with soft fruit using state of the art imitation learning methods infused with tactile feedback, as well as providing a simple but effective approach to damage assessment for post-harvest quality control. \newline}

\noindent Our key contributions are as follows:
\begin{itemize}
    \item We demonstrate the integration of optical tactile sensing into diffusion-based imitation learning policies for soft body manipulation.
    \item We demonstrate quantitative and qualitative reductions in bruising and improvements in success rate compared to our baselines while using tactile sensing. \review{We show fruit remains damage free after up to 50 repeated grasps.}
    \item We introduce FruitSplat, a novel method for qualitative and quantitative damage analysis for strawberries. 
    
\end{itemize}

\section{Related Work}
\label{sec:related_work}
\subsection{\review{Robotic Fruit Manipulation}}
\review{Robotic fruit manipulation is still relatively underexplored in literature, with most applications targeting harvesting as opposed to the post-harvest processes. Previous works include robotic fruit picking of apples, strawberries, blackberries, and peppers \cite{lv2019method, yu2019fruit, qiu2023tendon, hohimer2019design, de2024compliant, junge2023lab2field}.} These works typically involve the construction of a novel soft gripper, which is unique to a specific fruit or berry. In \cite{lv2019method} they focus on apples, and \cite{qiu2023tendon, de2024compliant} on blackberries. While these grippers generally perform well, they are limited to a single type of fruit. They typically prevent damage through modality specific designs \cite{de2024compliant, hohimer2019design, porichis2024imitation} and low dimensional sensing \cite{junge2023lab2field} or by not touching the fruit at all \cite{kim2025autonomous}. In this work, we focus on a general gripper capable of manipulating arbitrary fruit. \review{Rather than relying solely on soft robotics techniques or complex designs, we utilize optical tactile sensing.}

\review{Some recent studies have explored the utility of optical tactile sensing in agricultural applications, particularly for fruit manipulation and evaluation.} For instance, \cite{han2024learning} employs optical tactile sensing to facilitate fruit manipulation; however, their approach is limited to hard fruits such as apples and lemons, which restricts its applicability across a diverse range of produce. Furthermore, they do not quantify the damage over repeated grasping. In \cite{he2024cherry} the authors utilize GelSight \cite{donlon2018gelslim}, a vision-based tactile sensor, to determine the stiffness of cherry tomatoes. While our current work does not incorporate direct stiffness measurements, the underlying principles of this approach could be integrated into our pipeline. Furthermore, we focus on the manipulation task while preventing damage.

\vspace{-1mm}
\subsection{Bruise 
and Metrics}
Current approaches to bruise detection largely rely on qualitative assessments, with little emphasis on quantitative metrics. 
For example, \cite{junge2023lab2field} employs a visual classification system for raspberries, classifying them as undamaged, mildly damaged, or damaged, which suffers from subjectivity and limited resolution in damage assessment. As quality control of fruit on the market is commonly conducted through visual inspection, allowable fruit quality varies based on inspector \cite{opara2014bruise}. This system is inherently subjective and is in need of a standardized method to quantify damage.

There are a number of quantitative methods for damage detection for agricultural products. \cite{he2022recent} provides a comprehensive survey of high-precision methods currently in use. Techniques such as spectroscopy, optical coherence tomography (OCT), magnetic resonance imaging (MRI), and x-ray imaging offer detailed insights into 3D tissue damage, yet their reliance on expensive and complex setups restricts their practical use in robotics. While these techniques offer high resolution and rich information about the quality of a fruit, this is often unnecessary to determine if a fruit should be brought to market. \review{Multi-spectral imaging has also been used \cite{he2022recent}, however this can also require specialized hardware. A broad body of work has quantified fruit damage using stiffness-based metrics using direct contact force measurements \cite{moggia2017firmness, yang2022internal}. While informative, these metrics generally fall short of providing fine-grained analyses.}

Deep learning has also been explored to overcome some of these limitations. For example, \cite{unal2024detection} demonstrated bruise detection on red apples using convolutional models, while \cite{zhou2022deep} addressed strawberry bruise detection aided by UV light. Although these methods enhance detection capabilities compared to purely visual grading, they do not fully leverage the three-dimensional metric information of the fruit.

\vspace{-1mm}
\subsection{\review{Imitation Learning and Tactile Sensing}}

Recent advances have significantly improved the performance of  imitation learning through the use of Denoising Diffusion Probabilistic Models (DDPM)  \cite{chi2023diffusion}. These approaches learn observation conditioned, high-dimensional policies with relatively few demonstrations. \review{In \cite{mahmoudi2024leveraging} the authors survey the landscape of imitation learning for agriculture. While there are many examples of agricultural tasks like peeling bananas and identifying fruit, none rely on tactile sensing or use Diffusion Policies. Simple tactile sensing through force conditioning is used in \cite{huang20243dvitac} to pick up fragile grapes and placing them on a plate. } However, DexFruit not only safely grasps diverse fragile fruit using expressive tactile sensing, but also \textit{quantitatively} measures the damage done to the fruit across hundreds of trials providing a rigorous and necessary analysis of the fruit after manipulation. 

Recently, several studies have explored incorporating tactile sensing into imitation learning policies, utilizing diverse modalities such as audio \cite{liu2024maniwav}, low-resolution capacitive sensing \review{\cite{huang20243dvitac,wistreich2025dexskin}}, \review{and optical tactile sensing \cite{jiang2025gelfusionenhancingroboticmanipulation, xue2025reactive}}. \review{Reactivate Diffusion Policy (RDP) \cite{xue2025reactive} delivers impressive dexterity (e.g., cucumber peeling) via a two-tier tactile scheme, yet reduces tactile input to an optical deformation field. DexSkin \cite{wistreich2025dexskin} completes blueberry manipulation tasks with low-resolution capacitive sensing and reports no rigorous quantitative damage analysis; however, their qualitative examples show visible damage in over 40\% of fruit after a single move, whereas in our work we manipulate the same piece of fruit up to 50 times without significant damage}. Although these studies demonstrate impressive performance, our work is the first to rigorously validate tactile-feedback-based manipulation of delicate, soft fruits over hundreds of experimental trials.

\vspace{-1.0mm}
\section{Preliminaries}
\label{sec:preliminaries}
\subsection{Diffusion Policies}
Given an observation $\mathcal{O}$, which corresponds to external camera images and robot state, and an action $A$ in the space of robot positions, we seek to learn a policy $\pi: \mathcal{O} \rightarrow \mathcal{P}(A)$. Instead of learning this policy directly, we model our policy as a Denoising Diffusion Implicit Model (DDIM) and instead learn a noise predictor network following \cite{chi2023diffusion}: 
\begin{equation}
    \hat{\varepsilon}_k = \varepsilon_{\theta}(A^k_t, \mathcal{O}_t, k) 
\end{equation}
Where $A^k_t$ is a sequence of noisy actions, $\mathcal{O}_t$ is the observation and denoising iteration $k$. Given a dataset $\mathcal{D} = \{\{O_t, A_t \}\}_{t = 1}^n$ we train this denoising function by applying noise $\epsilon^k$ to data $A_t$ and minimizing the following loss:
\begin{equation}
    \mathcal{L} = \it{MSE}\Bigl(\epsilon^k, \varepsilon_{\theta}(A^k_t, \mathcal{O}_t, k) \Bigl)
\end{equation}
During training we sample random actions $A_t^K$ and perform denoising using a DDIM scheduler \cite{Song2021}.
\vspace{-1.0mm}
\subsection{3D Gaussian Splatting}
3D Gaussian Splatting (3DGS) is an explicit 3D visual representation for robotics. Given a series of 100-200 images, 3DGS constructs a visually accurate representation that can be rendered at high FPS on a GPU. \review{3DGS utilizes a set of 3D Gaussian primitives $g_i$ with learnable parameters, including 3D mean position $\mu \in \mathbb{R}^3$, 3D covariance matrix $\Sigma \in \mathbb{R}^{3x3}$, opacity $\alpha \in \mathbb{R}$, and spherical harmonic (SH) coefficients to represent color.
Each Gaussian can be defined as:
\begin{equation}
    \label{eqn:gaussian}
    G(x) = e^{-\frac{1}{2}(x)^T\Sigma^{-1}(x)}
\end{equation}
The 3D scene gets initialized typically from a sparse point cloud, which is computed via Structure-from-Motion (SfM) \cite{7780814}. Images are rendered by projecting the 3D Gaussians onto the image plane using an affine approximation of projective transformation.}
\noindent\review{The final pixel color $C$ is computed as follows:}
\vspace{-2mm}
\begin{equation}
    C = \sum_{i \in N} c_i \alpha_i \prod_{j=1}^{i-1} (1 - \alpha_j)
    \label{eqn:3dgs_march}
\end{equation}
where $\alpha_i$ is given by evaluating a 2D projection of a Gaussian with covariance $\Sigma$. The loss used to supervise the Gaussian Splat is a weighted combination of the L1 and \review{Structural Dissimilarity Index Measure (D-SSIM) loss} between each ground truth (GT) image and volumetric rendered image.
\begin{equation}
\mathcal{L} = (1 - \lambda)\,\mathcal{L}_1 \;+\; \lambda\,\mathcal{L}_{D-SSIM}
\label{eqn:3dgs_loss}
\end{equation}

\section{Method}
\label{sec:method}

DexFruit combines a robust manipulation policy and advanced 3D vision techniques to address the challenges of handling delicate fruits in agricultural robotics. We utilize Diffusion Policy trained on expert demonstrations with optical tactile sensing. To quantify fruit damage, we develop FruitSplat, a novel 3D Gaussian Splatting-based analysis tool that accurately reconstructs and compares fruit models before and after manipulation to quantify damage caused during manipulation.

\subsection{Fruit Manipulation Policy}

Safely handling fruit in agricultural robotic manipulation requires policies that balance two critical aspects: robustness, to reliably pick and deposit fruits, and sensitivity, to prevent damage during handling. These factors inherently conflict, as robustness encourages a firm grip, whereas damage prevention demands minimal contact. To meet these conflicting demands, we leverage a Diffusion Policy (see Section \ref{sec:preliminaries}), known for smooth and safe execution in complex real-world scenarios. While many existing approaches achieve gentleness through soft gripper hardware, our standard parallel-jaw gripper equipped with DenseTact implicitly achieves softness via the learned policy, successfully minimizing fruit damage.

For this task, our action space consists of the end-effector pose (3D position + 6D rotation \cite{zhou2019continuity}) and the width of the gripper normalized between [0,1]. The observation space consists of one RGB camera mounted on the robot arm wrist and RGB images from two DenseTacts. We perform data collection and inference at approximately 10hz. To accentuate the changes in the gripper state during contact, the average channel image difference is computed for each sensor. In addition, we include the current robot state as an observation in the same format as the action. \review{The RGB images are down-sampled to (224x224) and encoded into 512-dimensional latent vectors using ResNet-34 \cite{he2016deep}, while the DenseTact images are processed with ResNet-18 \cite{he2016deep}. Both ResNets are trained from scratch and implemented from PyTorch Image Models (TIMM) \cite{rw2019timm}. We train the model for 1000 epochs using a DDIM noise scheduler \cite{Song2021} with 50 train steps and 16 inference steps. We train with a batch size of 128 and a learning rate of 3e-4 across 4 L40 GPUs}. Expert demonstrations are collected via teleoperation, varying the robot's starting position and individual fruit to ensure the policy conditions on the fruit itself. To ensure consistency across methods and enable a fair comparison with a purely tactile sensing baseline, we fix the initial location of the strawberry. Since diffusion policies have already shown strong capabilities in identifying and grasping objects from diverse positions, we focus exclusively on gentle grasping rather than localization in this work.

During testing, we observed that raw DenseTact images alone provided insufficient signal strength for the policy to effectively leverage tactile information. This was due to minimal variation in the raw images upon contact, making learning challenging. We compute the image difference relative to the undeformed sensor and find that the resulting grayscale image contains sufficient information. Additionally, we noticed that straightforwardly combining vision and tactile images led to worse performance than using either modality individually, likely caused by the model overfitting to the visual inputs. Inspired by human tactile exploration, we resolved this issue by adaptively switching off visual input when tactile contact occurs, which proved essential for high-performing policies. Contact detection is implemented by computing the sum of squared pixel differences (after converting to grayscale), and triggering the visual–tactile switch when this energy exceeds a predefined threshold $\tau$.

\begin{figure}[t]
    \centering
    \vspace{.5mm}
    \includegraphics[width=0.99\linewidth]{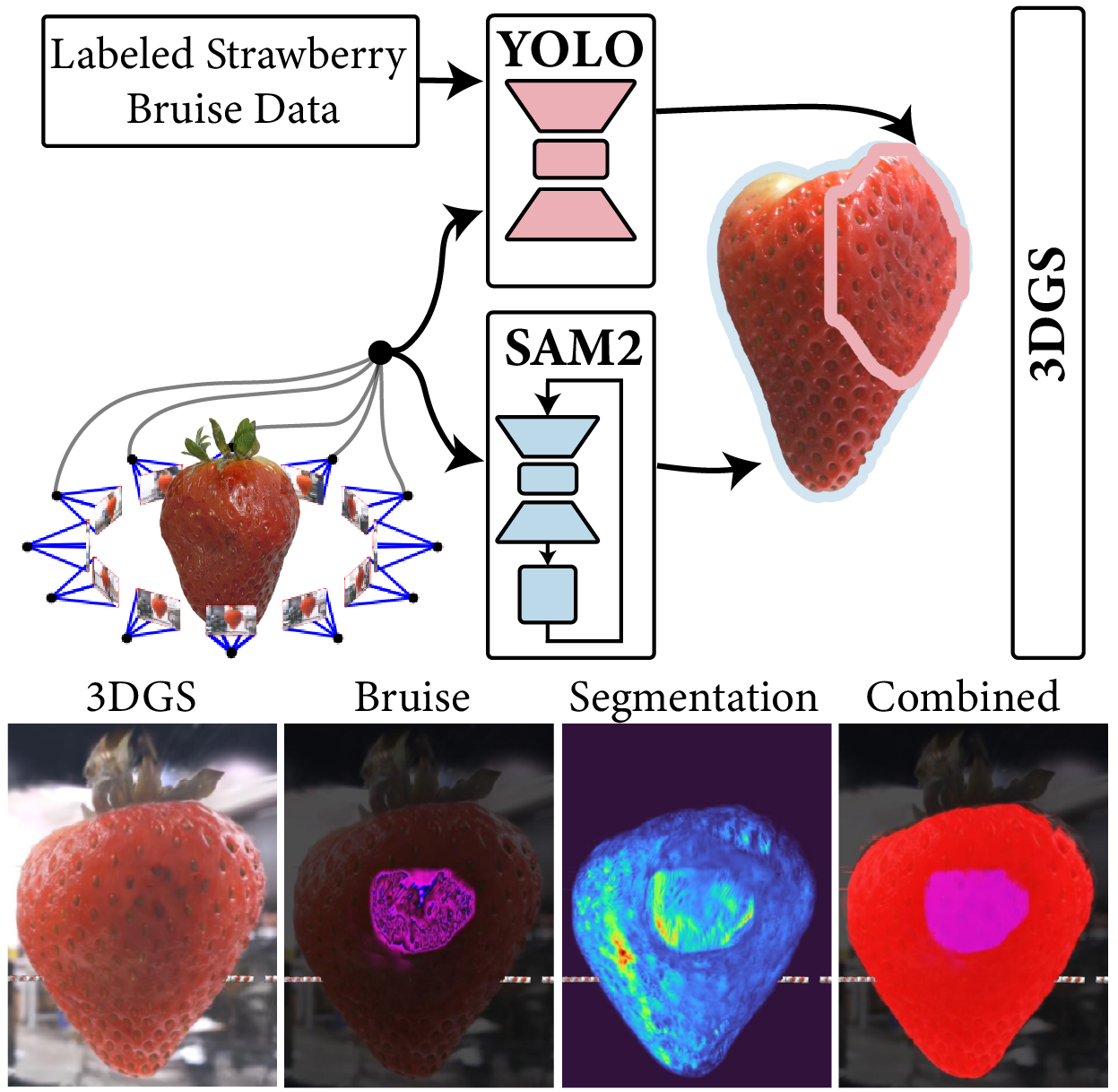}
    \caption{Here we show the FruitSplat pipeline. Prior to training, the 3D Gaussian Splat the images are processed through 2 models. YOLO segments the bruising, while SAM2 segments the strawberry. These two additional masks are used to supervise additional Gaussian parameters.}
    \label{fig:fruit_splat}
\end{figure}

\subsection{FruitSplat}

\review{For this work, we chose to implement FruitSplat only for strawberries and discuss the reasons for this decision in Section \ref{sec:limitations}.} To quantify damage made to strawberries, we leverage 3D Gaussian Splatting to accurately construct strawberries models before and after manipulation. FruitSplat has the following pipeline: 

\subsubsection{Data Pre-Processing} \review{Videos of the entire target strawberry are first collected using a web-cam mounted on a turntable. To ensure good coverage of the entire surface the strawberry is suspended on two pieces of fishing line as shown in Fig. \ref{fig:gripper}}. These images are then fed directly through COLMAP \cite{schoenberger2016mvs} to obtain sparse points and camera poses of each frame. We then pass each frame into our custom fine-tuned YOLOv12 \cite{tian2025yolov12attentioncentricrealtimeobject} bruising segmentation model for predicted bruise masks. \review{We fine-tuned the model on two thousand hand-labeled images of store-bought strawberries, with bruise masks manually annotated using open-source data labeling tools. The final segmentation model utilized in experiments achieved 94.2\% precision, 89.5\% recall, and 84.7\% Intersection Over Union (IoU) for our evaluation dataset of 162 images.} We also apply SAM 2 \cite{ravi2024sam2} to retrieve segmented strawberry masks.

\subsubsection{3D Gaussian Splatting} Our custom method directly inherits from Nerfstudio's Splatfacto \cite{Tancik2023}, \review{however we augment the Gaussian representation by adding additional parameters $s_i$ and $b_i$ to each Gaussian primitive \cite{shorinwa2024fast}. The parameters $s_i \in [0, 1]$ and $b_i \in [0, 1]$ are probabilities that a Gaussian $g_i$ is part of a strawberry and in a bruised region. Strawberry and bruise mask ground-truth maps can be provided as 2D images for supervision. $b_i$ and $s_i$ are rasterized similarly to color \eqref{eqn:3dgs_march} and are supervised using a binary cross entropy loss.}
Our dataloader utilizes the pre-processed YOLO and SAM2 masks as pseudo-ground truth for the supervised strawberry and bruise mask learning. During training, bruise and strawberry loss terms are simply added to the main loss. \review{Additionally, we gate the bruise losses on the strawberry mask as bruising should only occur on the strawberry and nowhere else in the scene. In order to improve training stability we use a stop-grad in the rasterization of bruise and strawberry masks, which only allows gradients with respect to the $s_i$ and $b_i$ parameters of each Gaussian and does not allow gradients to flow to position and shape parameters of the Gaussians.}

\subsubsection{Post-Processing and Damage Analysis} Once the Gaussian Splat is fully trained after 10,000 steps, FruitSplat exports the Gaussian means along with the learned Gaussian strawberry and Gaussian bruise parameters. It then self-filters the point cloud based on the strawberry parameter using a fixed probability threshold, leaving a high-resolution point cloud model of the strawberry. Each point is also exported with an encoded bruise score learned from the previously trained 3DGS, which represents the probability of being a bruise. FruitSplat then utilizes the per-point encoded bruise score to calculate the percentage of bruised points \review{within this filtered strawberry point cloud} that pass a preset confidence threshold, and subsequently compares the difference in bruising between pre-manipulation and post-manipulation strawberry. Due to the fact that the point cloud points already represent only the surface of the strawberry, this method can effectively represent visual surface bruising information. It should be noted that damage analysis requires a pre-manipulation and post-manipulation Gaussian Splat of the strawberry as an input; thus, steps 1 and 2 for FruitSplat must be run independently twice. However, both steps only require the use of a camera, with no extra hardware necessary.

\begin{figure}[t]
    \centering
    \vspace{.5mm}
    \includegraphics[width=0.99\linewidth]{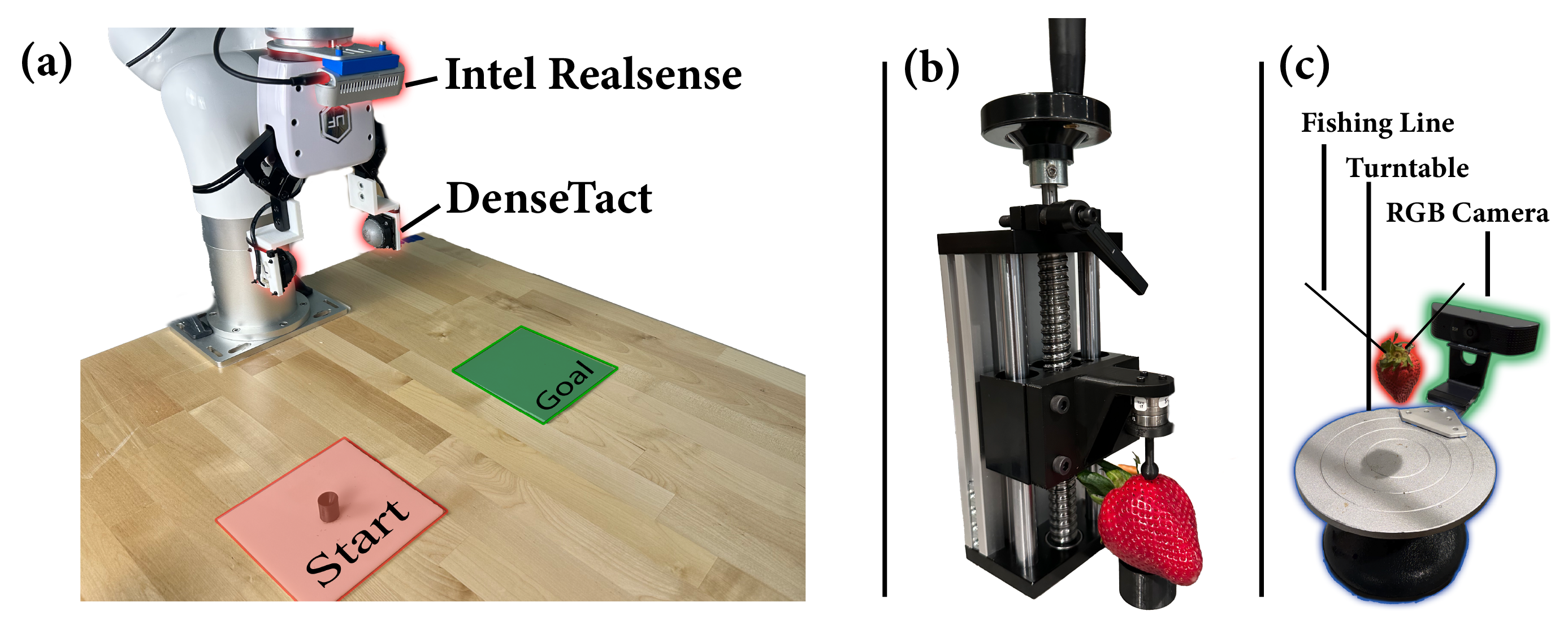}
    \caption{\review{On the left (a) is the experimental setup. The two sensing modalities used are highlighted in red: the DenseTact optical tactile sensor and IntelRealsense D435i. On the right (b) is the force measuring apparatus which is used to quantify internal damage in the fruit. (c) Shows that apparatus we utilize for scanning the strawberries.}}
    \label{fig:gripper}
\end{figure}

\review{\subsection{Internal Stiffness Metric}}

\review{As discussed in Section \ref{sec:related_work} fruit stiffness can be correlated with post harvest damage \cite{moggia2017firmness,yang2022internal}. }Furthermore during testing, we found that there were often instances where the internal structure of the fruit was significantly compromised but did not show observable visual changes. In order to measure this internal damage, we assess fruit damage before and after experimentation by measuring the fruits ability to resist an external force. We developed a system that measures the spring constant of the fruit surface under small deformations. In practice, we find that this accurately measures internal damage to the fruit. We calculate this metric by deforming the fruit a known distance and measuring the associated force required for the deformation. This setup can be seen in Fig. \ref{fig:gripper}. We select equal spaced points along the circumference of each fruit and position the fruit so that the point is normal to a 3D-printed circular indenter. We measure the force in the z-direction after a 1.25mm displacement of the z-axis linear guide. The resulting force is measured using an ATI F/T Sensor Nano17. By averaging the spring constant of the four points on each fruit, we compare the effectiveness of the gripper with the DenseTact and its baselines before and after gripping. This metric, \textit{\% Stiffness}, is the percent of pre-interaction stiffness, which is maintained after the experiment. $100\%$ is the best possible value. Blackberries are excluded from this metric due to their high susceptibility to damage.
\newline

\section{Results}
\label{sec:results}

We demonstrate significant improvements over our baselines in grasping success rate and damage reduction over 630 trials across strawberries, tomatoes and blackberries. \review{Strawberries are fragile, vary greatly in size, and show clear bruising on their surface after damage, making them an ideal challenge fruit for evaluating our method. Cherry tomatoes were selected due to their firm yet delicate skin, which makes them particularly sensitive to improper handling. Blackberries present additional complexity with their soft, easily ruptured structure, necessitating precise manipulation to avoid damage. Together, these fruits provide diverse tactile challenges critical for demonstrating the versatility of our approach.}

\begin{figure*}[t]
    \centering
    \vspace{.5mm}
    \includegraphics[width=0.95\linewidth]{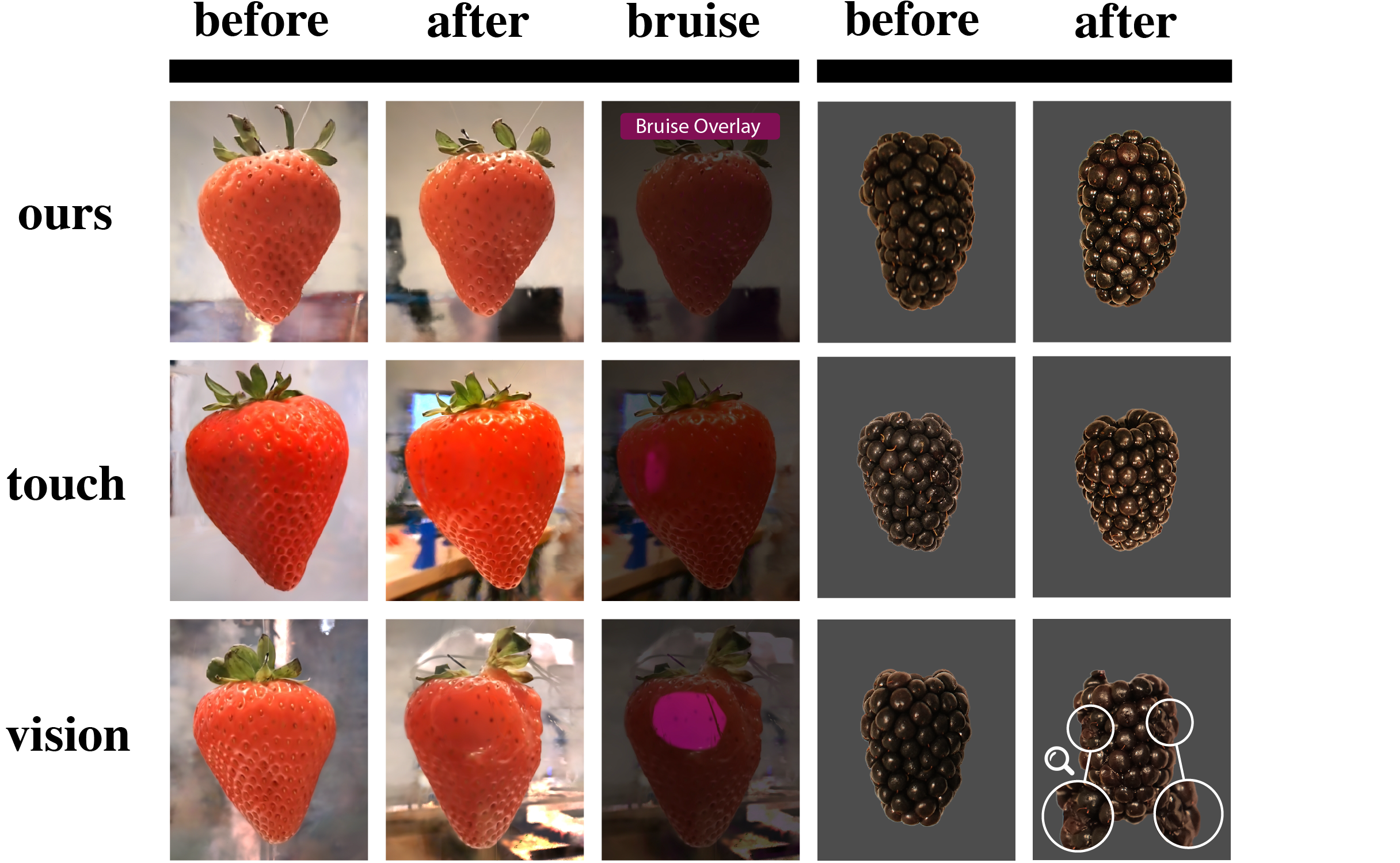}
    \caption{We show the qualitative results of our method compared to the baselines. On the left, we show the same strawberries before and after the experiments along with the corresponding bruise mask. All the strawberry images are rendered from FruitSplat. Blackberries are shown on the right while tomatoes are not shown as they do not show visible damage.}
    \label{fig:results}
\end{figure*}

\begin{table*}[htbp]
  \centering
  \renewcommand{\arraystretch}{1.25}%
  \resizebox{0.75\textwidth}{!}{%
    \begin{tabular}{|l|ccc|cc|c|}
      \hline
      \multirow{2}{*}{Method}
        & \multicolumn{3}{c|}{Strawberry}
        & \multicolumn{2}{c|}{Tomato}
        & Blackberry \\ \cline{2-7}
        & Success↑ 
        & Stiffness↑ 
        & Bruise ↓ 
        & Success↑ 
        & Stiffness↑ 
        & Success↑ \\ \hline
      \textbf{Ours}   & \textbf{95\%} & \textbf{80\%} & \textbf{0.00\%} 
             & \textbf{100\%} & \textbf{95\%} & \textbf{80\%} \\
      Touch  & 91\% & 78\% & \textbf{-1.75}\%
             & \textbf{100\%} & 93\% & 55\% \\
      Vision & 67\% & 29\% & 14.61\%
             & 69\%  & 77\% & 58\% \\ \hline
    \end{tabular}%
  }
  \caption{Experimental results. Bold indicates best with margin.}
  \label{tab:TabulatedResults}
\end{table*}

\subsection{Experimental Setup}

The experimental setup is shown in Fig. \ref{fig:gripper}. From a starting position, our policy is able to move to the strawberry and a grasp is attempted. The strawberry is then lifted and placed on the green square (goal). Any strawberry that is moved by the policy to the goal is labeled a success, otherwise it is labeled a failure. Failure modalities include missing a grasp of the strawberry completely or dropping the strawberry before reaching the target zone. Following best practices, the strawberries which are used for expert demonstrations are discarded and new strawberries are used for testing. We used the same two DenseTact sensors for the entire course of the project, across thousands of demos and experimental roll-outs.

To clearly highlight differences between the evaluated methods, we conduct a series of pick-and-place trials on strawberries, tomatoes, and blackberries. Specifically, we perform 25 trials per each strawberry, 50 per each tomato, and 20 per each blackberry. By conducting such a large number of trials, we both demonstrate the consistency of our method and create a key point of distinction from prior work. For each fruit type, we further vary the size categories: strawberries are tested across three size ranges (small: $<$4 cm, medium: 4–6 cm, large: $>$6 cm), tomatoes across two sizes (small: $<$3 cm, medium: $\geq$3 cm), and blackberries across three sizes (small: $<$2.5 cm, medium: 2.5–3.5 cm, large: $>$3.5 cm). These size categories are based on measured fruit length. \review{For each method and fruit size we perform all  trials on the \emph{same} fruit. This allows subtle damage to build up and become apparent, while showcasing the true delicacy of our method.} To ensure consistency, all fruit used in the experiment were sourced on the same day from a single lot, ensuring similar ripeness and quality. The fruit starts on a small 3D-printed pedestal positioned in the center of the white start square. The gripper starts 20 cm away from the strawberry and moves on average 20 cm to the goal region. We employ an Intel Realsense D435i camera positioned on the robot wrist, angled toward the gripper. We utilize the Ufactory xArm7 operating in position control mode for the experiments. 

Data collection is performed via a 3D Connexion space mouse and a mouse-based slider to control gripper position. Additionally, for our full method with tactile sensing, the RGB images of the DenseTact sensors are displayed to the teleoperator. We collect between 100 and 200 demos for each fruit type, ensuring a wide range of variability in our dataset. As our baselines are ablations of the full method, we use the same set of demos for each. We take 360-degree videos of each target fruit before and after the manipulations trials. These videos are then fed into our FruitSplat pipeline to produce quantitative damage results.

\subsection{Baselines}
\label{sub:baselines}
In order to effectively evaluate our method, we compare our results to a number of baselines and ablations. \textit{Vision} is our full gripper setup with the DenseTact sensors except the model does not incorporate touch information and instead relies only on the wrist-mounted camera. Likewise \textit{Touch} relies only on the tactile input. Finally we also test a naive combination of vision and touch where both inputs are combined without any natural switching labeled \textit{Vision+Touch*}.

\subsection{Experimental Results}

\begin{figure}
    \centering
    \includegraphics[width=1.0\linewidth]{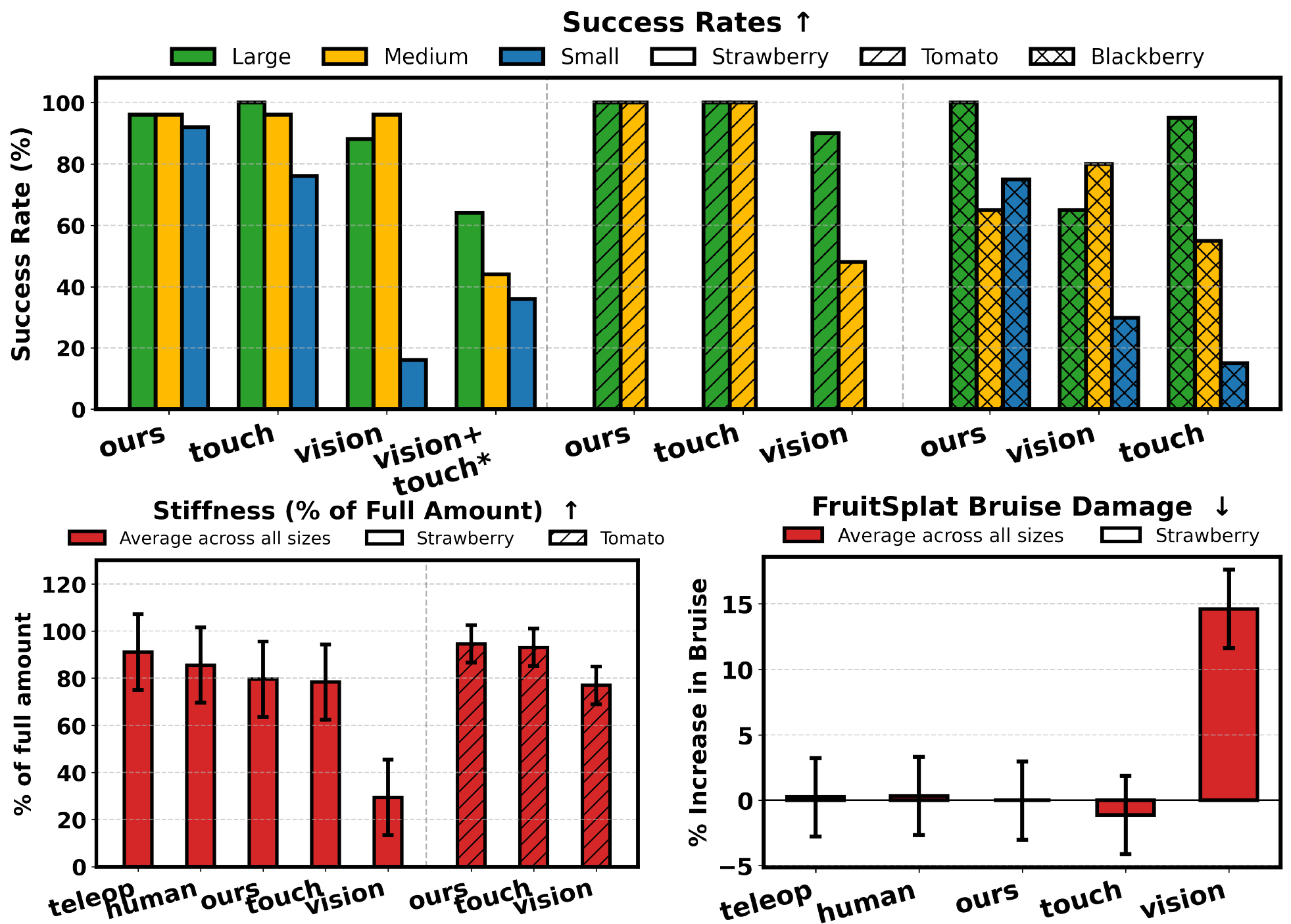}
    \caption{Above we show the success rates across methods and fruit size. Below on the left we quantify the internal fruit damage measured as a percent change in stiffness, while on the right we show the results from FruitSplat representing external damage.}
    \label{fig:results_chart}

\end{figure}

We perform a rigorous experimental testing program to effectively compare our method to the baselines of vision-only and touch-only. The quantitative results are shown in Tab. \ref{tab:TabulatedResults} and Fig. \ref{fig:results_chart}, while the qualitative images are shown in Fig. \ref{fig:results}. To emphasize the differences between methods we run the same fruit with multiple grasps, 25 for strawberries, 50 for cherry tomatoes and 20 for blackberries. 
Examining the results in Tab. \ref{tab:TabulatedResults}, we see that our method outperforms or matches the baselines in success rate across all three of the tested fruits, while also achieving the least damage across our measured metrics. Success is defined as moving the fruit from the start to the goal and is independent from damage. Specifically for strawberries, we found that post-manipulation bruise difference was almost zero, compared to the worst performing baseline of vision only with 14.6\% bruising. It should also be noted that the negative percentage in strawberry bruising for touch only is likely caused by slight inconsistencies in pseudo-ground truth bruise mask detection by our YOLO model. We refer the readers to Fig. \ref{fig:results_chart} which includes our estimated error bars based on the variance across trials. 

Fig. \ref{fig:results_chart} provides more detail showing the success rate across the different fruit sizes. We can see that the vision only methods struggle the most on small and medium fruit. We hypothesize that the vision only failure is due to an overfitting on size where the model fits to the average fruit size. While this leads to good grasping success rate on large fruit, it causes grasps which are far too loose on small and medium fruit. Furthermore, this often leads to over-squeezing of larger sized fruit causing more damage. This increased damage is clearly visible on the bottom of Fig. \ref{fig:results_chart}. Here we show the \% change in stiffness relative to before the experiment. For comparison, we also include both \textit{human} where a human performs the same action as the robot for each trial, and \textit{teleop} where a human teleoperator controls the robot. While \textit{ours} and \textit{touch} perform similar to the theoretical optimal of \textit{teleop} and \textit{human} the \textit{vision} policy causes a much greater amount of damage. Furthermore, we see that the vision only policy leads to much more external bruising. 

We can also see in Fig. \ref{fig:results_chart} an additional ablation labeled \textit{vision+touch*} which combines vision and touch without any intelligent switching. As discussed in Section \ref{sub:baselines} this policy performs poorly. We speculate this is due to the policy ignoring the latent tactile vector and overfitting on visual information even after contact has occurred. We can see that unlike the vision or touch only policies, this policy is worse across all three sizes of fruit indicating that it suffers from a different mode of failure. 

Finally Fig. \ref{fig:results} shows the qualitative results of our experiments. We can see clear differences in bruising across our method and baselines. The high quality visual accuracy of the FruitSplat reconstructions of the strawberries are also visible. Tomatoes are not shown as they exhibit nearly no visible bruising when damaged. 

\section{Conclusion}
\label{sec:conclusion}
This work presents DexFruit, an optical tactile sensing embedded diffusion policy for soft fruit manipulation. We demonstrate that an intelligent switching between visual and tactile inputs greatly improves soft handling of fragile fruits such as strawberries, tomatoes, and blackberries compared to utilizing pure vision or touch sensing. We also present FruitSplat, a novel technique that harnesses the geometric accuracy and photorealism of 3D Gaussian Splatting to qualitatively and quantitatively represent fruit damage post-robot manipulation. Experimentation across various policy ablations and fruit types demonstrate DexFruit's  best performance in both policy grasping success rate and post-manipulation damage minimization, demonstrating an average 92\% grasping policy success rate, up to a 14.6\% reduction in visual bruising.

\section{Limitations and Future Work}
\label{sec:limitations}
While DexFruit presents improvements in dexterous manipulation of soft fruits, some limitations and several opportunities for future work exist. \review{With regards to our fruit selection for FruitSplat, we chose to restrict our initial scope of work on strawberries chiefly due to the data-collection and training time associated with implementing more fruits, which is an inherent limitation of our method. In principle, we expect our method can extend to any fruit that displays visible bruising such as blackberries. This is also the reason we did not implement tomatoes, due to the absence of visible bruising. In our experience with manipulating tomatoes for the baseline methods, we found that the best way to quantify any damage was through our stiffness metric. Since the body of the tomatoes are substantially less stiff than their outer layer of skin, excessive gripping of tomatoes only resulted in internal damage and was visually undetectable. For future work, we hope to extend FruitSplat to other fruits through curating an extensive database of labeled fruit damage. Additionally, integrating multi-spectral data \cite{sinha2024spectralgaussians} into FruitSplat could further increase its utility and robustness.}
Our experiments were conducted in controlled laboratory settings rather than actual field conditions. \review{Further method evaluation should be done for both the harvesting and post-harvest processes out in the field and in produce warehouses.} We believe that integrated ripeness sensing could be achieved by leveraging soft optical tactile sensing output. This system could simultaneously improve yield by reducing damage and improving harvest ripeness.
On the policy side, the method for switching off vision and turning on tactile in our method is conditioned on a contact heuristic. While this is sufficient for this work, this could be extended to a learned mode switch as the policy chooses actively which input is more useful.

\bibliographystyle{IEEEtran} %
\bibliography{references} %

\end{document}